\documentclass{amia}
\usepackage{graphicx}
\usepackage[labelfont=bf]{caption}
\usepackage[superscript,nomove]{cite}
\usepackage[dvipsnames]{xcolor}
\usepackage[sort,comma,numbers,super]{natbib}
\usepackage{multirow}
\usepackage{float}
\usepackage{url}
\usepackage{adjustbox}
\usepackage{tabularx}
\usepackage{soul}
\usepackage{subcaption}  
\usepackage{ulem} 
\usepackage{tabularx}
\usepackage{graphicx}
\usepackage{hyperref}
\usepackage{listings}
\usepackage{amsmath}
\usepackage{needspace}
\usepackage{enumitem}

\usepackage{booktabs}
\usepackage{multirow}
\usepackage{makecell}
\usepackage{array}
\usepackage{pifont}
\usepackage{bm}
\usepackage{relsize}
\usepackage{amsmath}
\usepackage{subcaption}
\usepackage{sidecap}
\usepackage{fancyhdr}
\usepackage{csquotes}
\newcommand\EatDot[1]{}

\definecolor{myblue}{HTML}{4E79A7}
\definecolor{mygreen}{HTML}{59A14F}
\definecolor{myred}{HTML}{E15759}

\begin{document}

\title{Using Large Language Models to Generate Clinical Trial Tables and Figures }

\author{Yumeng Yang, MS$^1$ $^2$, Peter Krusche, PhD$^3$, Kristyn Pantoja, PhD$^4$, Cheng Shi, PhD$^1$, Ethan Ludmir, MD$^5$, Kirk Roberts, PhD$^2$, Gen Zhu, PhD$^1$\textsuperscript{*}}

\institutes{
    $^1$Novartis Pharmaceuticals Corporation, East Hanover, NJ, USA\\
    $^2$School of Biomedical Informatics\\ The University of Texas Health Science Center at Houston, Houston, TX, USA \\
    $^3$Novartis Pharma AG, Basel, Switzerland\\
    $^4$Novartis Pharmaceuticals Corporation, Cambridge, MA, USA\\
    $^5$Department of Radiation Oncology\\ The University of Texas MD Anderson Cancer Center, Houston, TX, USA\\
}
\maketitle

\begin{flushleft}
    {*}Correspondence to: Gen Zhu, \texttt{gen.zhu@novartis.com}
\end{flushleft}

\section*{Abstract}
\label{Abstarct}
Tables, figures, and listings (TFLs) are essential tools for summarizing clinical trial data. Creation of TFLs for reporting activities is often a time-consuming task encountered routinely during the execution of clinical trials. This study explored the use of large language models (LLMs) to automate the generation of TFLs through prompt engineering and few-shot transfer learning. Using public clinical trial data in ADaM format, our results demonstrated that LLMs can efficiently generate TFLs with prompt instructions, showcasing their potential in this domain. Furthermore, we developed a conservational agent named ``Clinical Trial TFL Generation Agent'': An app that matches user queries to predefined prompts that produce customized programs to generate specific pre-defined TFLs. 

\section{Introduction}
\label{Introduction}
In the pharmaceutical industry, submission of a clinical study report (CSR) is part of the drug approval process with health authorities. CSRs are highly standardized: For example, ICH (International Council for Harmonisation of Technical Requirements for Pharmaceuticals for Human Use) provides a guideline about the structure and content of the CSR (E3 Structure and Content of Clinical Study Reports | FDA). In the CSR, tables, figures and listings (TFLs) are the essential elements for summarizing clinical trial data, including, for example, summaries of demographic data, efficacy and safety results. Importantly, all analyses and TFLs that summarize outcomes are pre-specified from the protocol and statistical analysis plan through a set of highly detailed documents that are produced before any data are collected. Data shown in each TFL is pre-determined and traceable to data collection, and the layout of TFLs is specified via TFL shells – mockup versions of the table or figure that explicitly determine formatting, labels and footnotes. Once data have been collected, the industry practice to populate the TFLs is based on the data in  Study Data Tabulation Model (SDTM )  and Analysis Data Model (ADaM) formats which are introduced in 2006 by the Clinical Data Interchange Standards Consortium (CDISC) as data standards for drug submissions (CDISC ADaM)\cite{cdisc_adam}. 

To implement TFLs, statisticians and statistical programmers need to prepare and validate statistical programs from a data \& variable specification/data dictionary and a TFL shell that specifies the layout and calculation of each table or figure or listing. Once these programs and outputs are validated, they can be included in the clinical study report (CSR). This workflow ensures consistency, accuracy, and comprehensibility of the trial data, ensuring effective and un-biased analysis and interpretation. It critically relies on skilled statisticians and programmers to produce each TFL output and is time-consuming due to the large number of outputs, as well as complexity introduced by study-specific pre-specifications, such as specific custom variable mappings, counts and visual representations. 
Recent advances in large language models (LLMs) have demonstrated substantial potential to accelerate applications that involve text generation, classification, and natural language understanding\cite{luo2022biogpt,yang2022large,zhu2021twitter}. Such LLMs have already shown power in many fields\cite{wu2023bloomberggpt,shin2020biomegatron,fan2024advanced}; their applications in clinical trials have been particularly promising, with LLMs being
employed to assist in various tasks, such as the classification of clinical trial eligibility criteria and the extraction of relevant medical information from unstructured data sources. Some research focuses on using language models to extract the eligibility criteria from clinical trial protocols\cite{yang2023text,yang2024exploring}, while others explore using these models to assess patients' eligibility from electric health record (EHR)\cite{juhn2020artificial,zeng2018natural}. As LLMs have also been successfully applied to link natural language specifications and program code – e.g. by generating programs or test cases\cite{jiang2024survey}, using LLMs for statistical programming has received much attention recently\cite{coello2024effectiveness,impact_ai_cs_education}. 

In our work, we investigated the use of LLMs for generating TFLs for clinical trial data analysis. We focused on table and figure generation (a listing can be treated as a table) from tabular data. Tabular data are different in structure from the text datasets that LLMs are typically used and trained with. Integrating LLMs into the TFLs generation from tabular data is not straightforward, as the ability to understand table structure and analyze tables using LLMs has not been explored as thoroughly as work with plain text\cite{zhang2024survey,lu2024large,sui2024table}. Different from plain text documents, tables are structured with complex interrelations between rows and columns. Previously, table reasoning tasks such as table-based question answering\cite{iyyer2017search}, table-based fact verification\cite{chen2019tabfact} and table-to-text\cite{wang2021tuta} were often tackled by pre-training or fine-tuning neural language models. Recent development of LLMs provide potential to solve these tasks with higher accuracy\cite{zhang2024survey,lu2024large,sui2024table}. Our TFL use case presented the additional challenge of table-to-table/figure tasks, which may require advanced data manipulation and data analysis. A critical aspect of working with tabular data is enabling the model to understand the relationships between rows and columns, which is essential for accurate data interpretation. Three primary LLM-based approaches from the literature may provide solutions to our problem: 1) fine-tuning open-source LLMs with labeled data; 2) prompting LLMs directly by taking advantage of LLMs’ reasoning ability; 3) developing an LLM-based agent that can be integrated with other computing tools such as Python\cite{lu2024large,zha2023tablegpt,zhang2023tablellama}. In our study, we only considered the second and third approaches. 

This work focused on leveraging LLM-based approaches to automate the generation of TFLs that are guided in ICH E3 (E3 Structure and Content of Clinical Study Reports | FDA). More specifically, we investigated the use of LLMs to reproduce the outputs from the CDISC pilot dataset, an open-source dataset for testing tools and methods for clinical trial reporting and analysis\cite{atorus_research_cdisc_pilot_replication}. We considered demographic summaries, baseline calculations, and efficacy  summaries at different endpoints. More details about the approaches and data are discussed in the Methods section. The Evaluation and Results section discusses the performance of our approach and introduces an app based on the outputs generated by AI that can serve as an example for how this approach may be operationalized. Limitations and future development ideas are discussed at the end. 

\section{Method}

\subsection{Data Description}
\label{section:Data Description}

The data in our study is from the CDISC Pilot replication Github repository which aims to reproduce the table outputs within the CDISC Pilot Project using the PHUSE Test Data Factory project’s data\cite{phuse_test_dataset_factory}. This dataset has been used to demonstrate new analytical methods and tools for clinical studies, for example, utilizing open-source software tools such as the R programming language\cite{atorus_research_cdisc_pilot_replication}.

We started with datasets in ADaM format and aimed to produce TFL outputs from these. The goals of starting with the ADaM data format include easier and more efficient data review, replication, and output generation. The homogeneity of column names in the ADaM format simplifies the expansion to various trials in the future. The data used to generate the table results include the following datasets adhering to CDISC standards: adsl, adadas, adcibc, and adnpix. These datasets provide a comprehensive overview of the clinical trial, covering various dimensions of patient information and assessments on a subject level. The adsl dataset gives subject-level data with one record per subject, capturing demographics, treatment arms, and key study-specific attributes. In addition, adlb, adadas, adcibc, and adnpix provide data on visit-based assessments, offering detailed insights into cognitive function, clinician impressions, and neuropsychiatric symptoms, respectively. Together, these datasets allow for an in-depth analysis of treatment efficacy and patient outcomes throughout the trial. In order to derive summaries for TLF outputs, we typically subset some of these datasets to a specific set of events, visits or variables, joined with adsl for subject-level summaries and then produced listings or counts.

Our method involved leveraging LLMs as agents to generate Python code based on specific prompts or tasks related to data analysis. While R and Python are both among the commonly-used open-source statistical programming languages in the pharmaceutical setting, LLMs have been shown to yield executable code more often for Python than for R\cite{buscemi2023comparative}. SAS is also a commonly used statistical programming language, however it is proprietary. The model interprets the user's input, understands the task requirements, and generates the corresponding Python code by calling the LLM API. Once the code is generated, it is automatically executed in a Python interpreter with access to the input datasets to produce the results. This approach streamlines the data analysis process and allows incremental experimentation with the outputs by generating code and output simultaneously. Being able to inspect both code and output helps us to ensure accuracy and transparency in the analysis. Notably, we also experimented with prompting without the intermediate code generation step, having the model populate, e.g., table shells directly. However, we found that the accuracy of the table results was too low to pursue this solution further. 

\begin{figure}[h]
  \centering
  \includegraphics[width=1\textwidth]{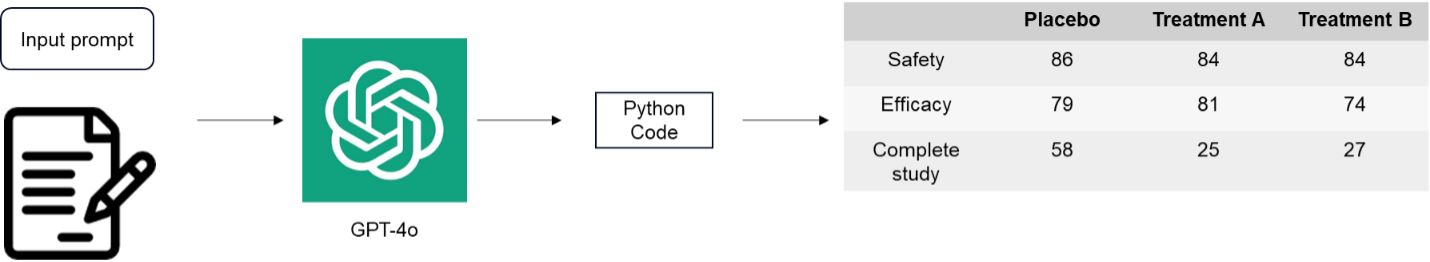}  
  \caption{Using the LLM as an Agent to Generate Results}
  \label{fig:your_label}
\end{figure}

\subsection{Prompts Design}

LLMs have been shown to be sensitive to the format of prompts\cite{zhao2021calibrate,lu2021fantastically,liu2024large}. In our approach, we provided three key components to the model: the system prompt, the user prompt, and a few-shot coding examples. The system prompt defines the model's scope of work; in our case, we set it as, “You are a statistical programming assistant,” to focus the model's responses on statistical analysis tasks. The user prompt contains detailed instructions to guide the model, including filtering relevant columns, categorizing data by treatment groups, performing counting calculations, and computing specific statistical measurements such as mean, median, standard deviation (SD), and range. More details on these instructions can be found in table 1.

Additionally, we included some Python code examples in the prompt to help the model understand how to read different formats of inputs files and convert byte strings to regular strings. We also instructed the model to avoid printing intermediate steps and instead directly generate the desired Python code, execute it automatically, and then display the final result. The first prompt was designed to generate Python code to calculate the summary statistics for each treatment arm, with the results shown in Figure 2. The second prompt was intended to generate Python code for plotting the Kaplan-Meier curve for time to the first dermatologic event, with the results displayed in Figure 3. Both sets of generated code matched the manually coded results with 100\% accuracy.

\begin{table}[H]
\centering
\begin{adjustbox}{max width=\textwidth}
\begin{tabular}{|p{5cm}|p{12cm}|}  
\hline
\centering\textbf{Prompt Example} & \textbf{Generated Python Code} \\
\hline
Please provide Python code to count the number of subjects in each treatment arm and then, within each treatment arm, count the number of subjects for safety (SAFFL), efficacy (EFFFL), Intent-To-Treat (ITTFL), Complete Week 24 (COMP24FL). The code should use \texttt{pd.read\_sas} with the \texttt{format} parameter set to \texttt{xport} to import the dataset file if it has an .xpt extension, otherwise use \texttt{pd.read\_csv} to import the dataset file if it has a .csv extension. &
\begin{lstlisting}[basicstyle=\footnotesize\ttfamily, language=Python]
import pandas as pd
# Load the dataset
df = pd.read_sas('adsl.xpt', format='xport')
# Convert byte strings to regular strings
for col in df.columns:
    if df[col].dtype == 'object':
        df[col] = df[col].apply(lambda x: x.decode('utf-8')
        if isinstance(x, bytes) else x)
# Count the number of subjects in each treatment arm
treatment_counts = df['ARM'].value_counts().reset_index()
treatment_counts.columns = ['Treatment Arm', 'Number of Subjects']
# Count the number of subjects for each flag within each treatment arm
flags = ['SAFFL', 'EFFFL', 'ITTFL', 'COMP24FL']
flag_counts = df.groupby('ARM')[flags].apply(lambda x:
x.apply(lambda y: (y == 'Y').sum())).reset_index()
# Merge the counts
result = pd.merge(treatment_counts, flag_counts, 
left_on='Treatment Arm', right_on='ARM').drop(columns='ARM')
# Print the result in table format
print(result.to_string(index=False))
\end{lstlisting} \\
\hline
You are a statistical programming assistant. You need to generate the code for a Kaplan-Meier plot for the time to the first dermatologic event, stratified by treatment arms, based on the data provided. The code should use \texttt{pd.read\_sas} with the \texttt{format} parameter set to \texttt{xport} to import the dataset file. Before processing the data, convert byte strings to regular strings. Here is some example code:
``
for col in df.columns:
    if df[col].dtype == object:
        df[col] = df[col].apply(lambda x: 
        x.decode('utf-8') if isinstance(x, bytes) else x)
''   
Use the entire dataset provided by the user without creating any sample data. Print the result in table format. Just give me the python code. &

\begin{lstlisting}[basicstyle=\footnotesize\ttfamily, language=Python]
import pandas as pd
import matplotlib.pyplot as plt
from lifelines import KaplanMeierFitter
# Load the dataset
df = pd.read_sas('adtte.xpt', format='xport')
# Convert byte strings to regular strings
for col in df.columns:
    if df[col].dtype == 'object':
        df[col] = df[col].apply(lambda x: x.decode('utf-8')
        if isinstance(x, bytes) else x)
# print the result in table format
print(df)
#create the Kaplan-Meier plot
kmf = KaplanMeierFitter()
#Plotting the Kaplan-Meier curve for each treatment arm
plt.figure(figzie=(10,6))

for treatment in df['TRAT'].unique():
    mask = df['TART']== treatment 
    kmf.fit(durations=df[mask]['AVAL'],
    event_observed=1-df[mask]['CNSR'],
    label = treatment)
    kmf.plot_survival_function()

plt.title('Kaplan-Meier Plot for Time to First Dermatological Event')
plt.xlabel('Time(days)')
plt.ylabel('Survival probability')
plt.legend(title='Treatment Arm')
plt.grod(True)
plot.show()
\end{lstlisting} \\
\hline
\end{tabular}
\end{adjustbox}
\caption{Prompt examples and codes generated by the model}
\end{table}

\begin{figure}[h]
  \centering
  \begin{subfigure}[b]{0.8\textwidth}  
    \centering
    \includegraphics[width=\textwidth]{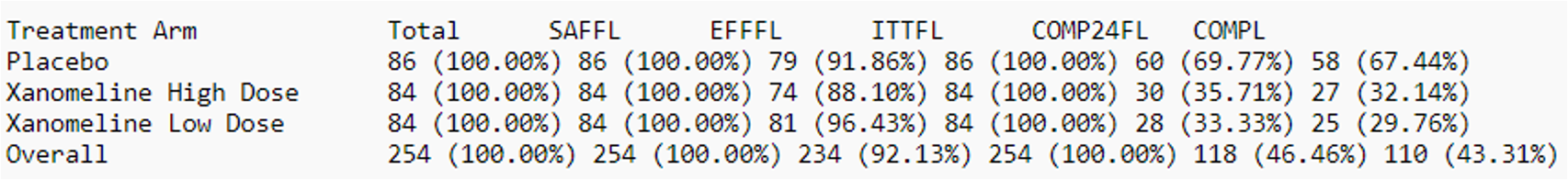}  
    \caption{Table result generated by prompt 1}
    \label{fig:table1}
  \end{subfigure}
  \vskip\baselineskip  
  \begin{subfigure}[b]{0.8\textwidth}  
    \centering
    \includegraphics[width=\textwidth]{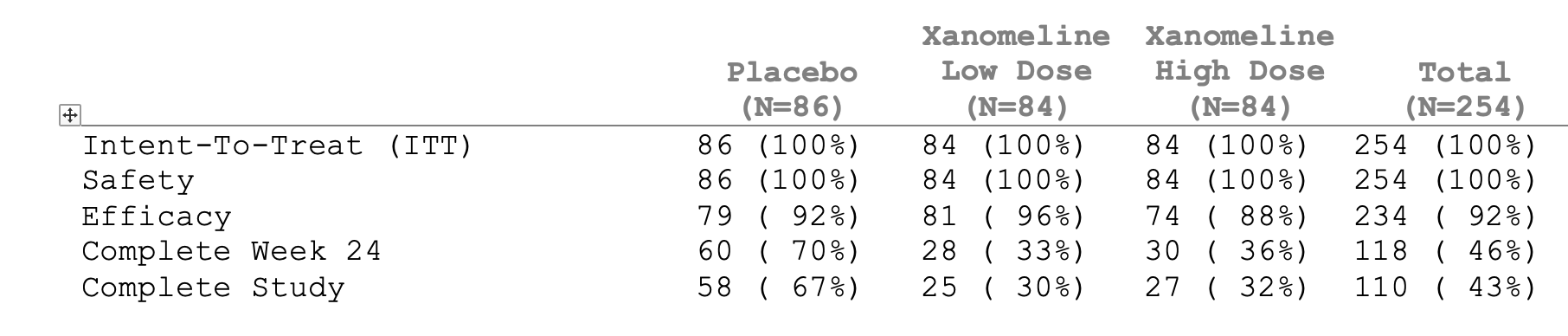}  
    \caption{Table generated by manually code}
    \label{fig:original_table1}
  \end{subfigure}
  \caption{Comparison of table results}
  \label{fig:comparison_tables}
\end{figure}

\begin{figure}[H]
  \centering
  \begin{subfigure}[b]{0.49\textwidth}
    \centering
    \includegraphics[height=5cm]{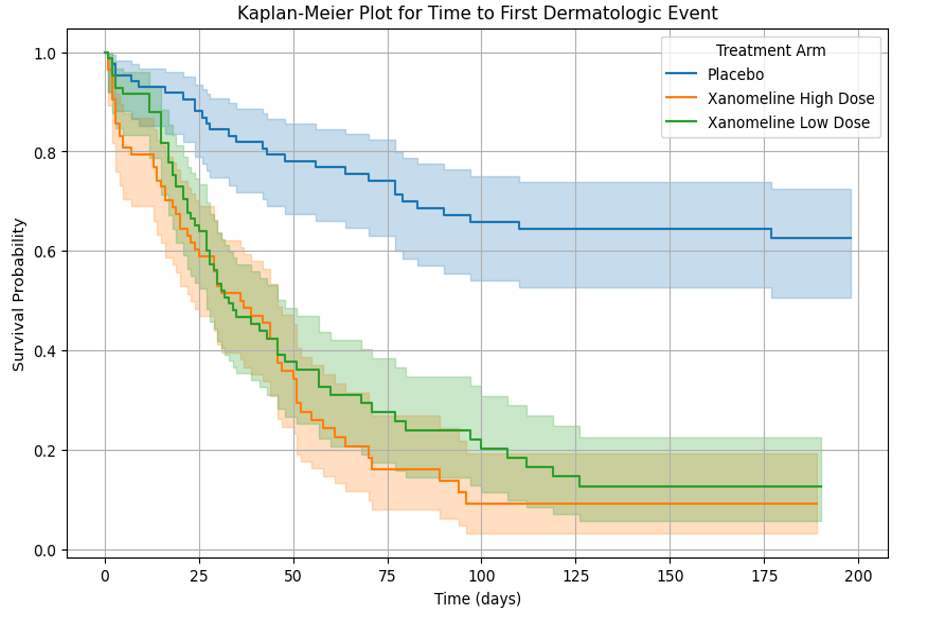}  
    \caption{Kaplan-Meier plot generated by prompt 2}
    \label{fig:kpplpt}
  \end{subfigure}
  \hfill
  \begin{subfigure}[b]{0.49\textwidth}
    \centering
    \includegraphics[height=5cm]{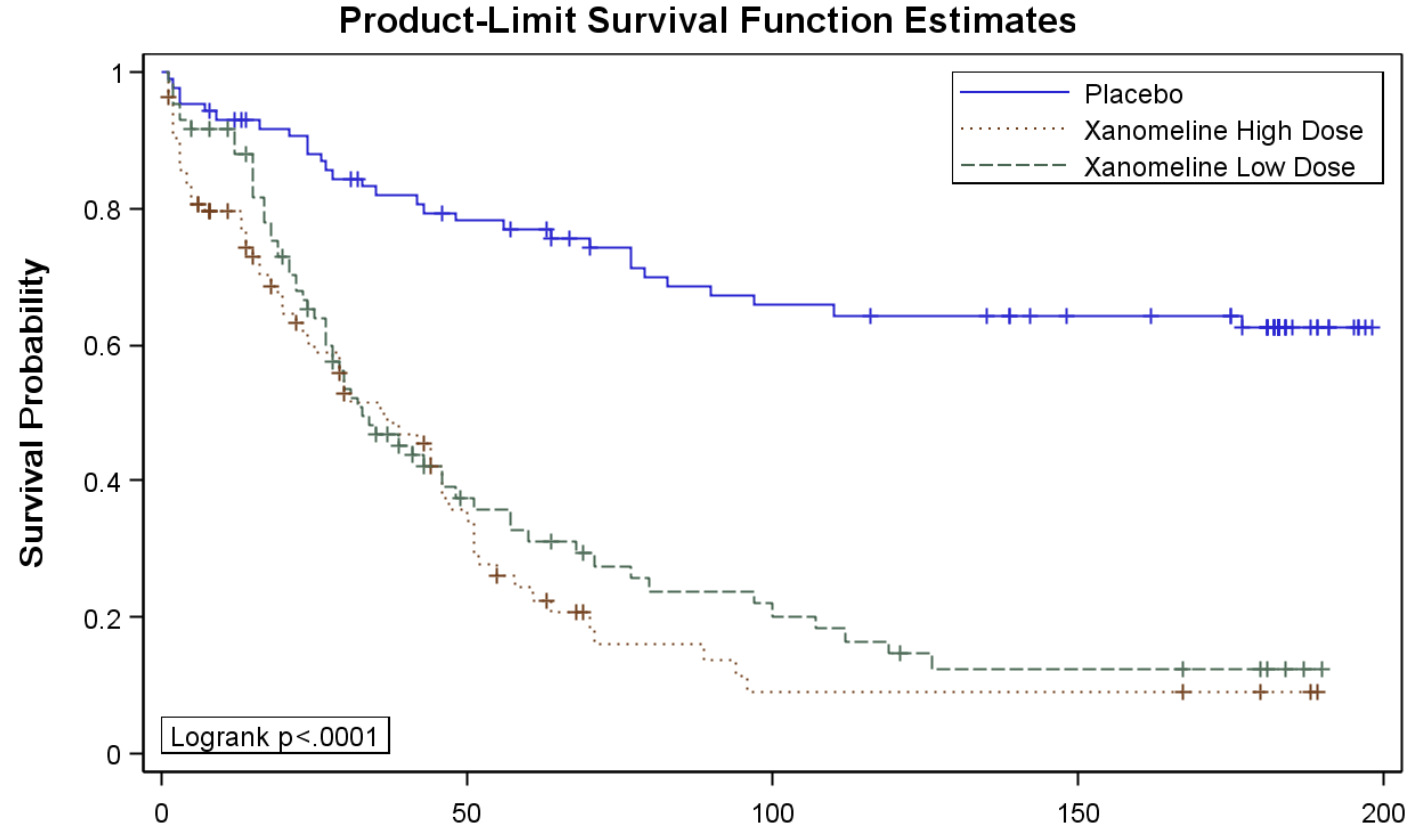}  
    \caption{Original Kaplan-Meier plot}
    \label{fig:original_kpplot}
  \end{subfigure}
  \caption{Comparison of Kaplan-Meier plots}
  \label{fig:comparison_kmplots}
\end{figure}

\section{Evaluation and Result}
Our work focused on generating 16 tables and 1 Kaplan-Meier plot. We manually compared all the table results generated by the LLMs with those published in the CDISC pilot replication project, and we also compared the generated Kaplan-Meier plot with the corresponding plot from the study report of the CDISC pilot study\cite{CDISC_sdtm_adam_pilot_project}. Our designed prompts achieved 100\% accuracy in replicating the results, except for analyses involving statistical tests, which need more prompt customization and input from statisticians to define the working steps for the model.

We then tested two predefined prompts for generating tables on a synthetic clinical trial dataset available at Novartis, using its adsl dataset. We attempted to generate two different tables where our prompts were generic enough to also cover a study in a different disease area (the previous study was centered on Alzheimer's disease trials, with some tables specifically tailored to its measurements). Although this new trial had slight differences from our previous one, we were able to reuse the existing prompts with only minor modifications to select the necessary columns; all other aspects remained unchanged. The prompts successfully returned the correct answers, matching the results we manually coded to generate the table outputs.

Our work further demonstrated the feasibility of using large language models like GPT-4o to generate table results from clinical trial ADaM datasets. The 16 prompts we have designed can accurately generate our desired tables. There are multiple endpoints in our example data, including week 8, week 16, and week 24, for each of these endpoint analyses, we use the almost the same prompt but just change the endpoint date, the model can easily identify the difference and return correct results.

\section{Dashboard Buildup}
The prompts we designed still require some domain and statistical knowledge. To study how we can make the prompts accessible to study team members who may not be familiar with LLMs and prompt design, we have developed our app which we call the Clinical Trial TFL Generation Agent. It is a chatbot-based dashboard that allows users to interactively generate specific clinical trial tables, utilizing a two-layer architecture to ensure precise and relevant responses. When a user makes a request, the model doesn’t attempt to guess their intent; instead, it matches the request to one of several predefined prompts, ensuring that the model accurately identifies the information needed and how to present it, regardless of how the request is phrased. 

The first layer, called the ``Table Category Description,'' focuses on understanding the user’s request by summarizing the overall purpose of the table, such as providing a population summary by different treatment arms. Once the request is clearly understood, the second layer, ``Stepwise Detailed Prompts'' is activated. This layer contains a fixed set of technical steps necessary to create the table, offering detailed instructions on data categorization, filtering relevant columns, and including Python code examples for tasks like reading different data formats. This structured approach ensures that the data analysis process is efficient, transparent, and easily traceable, as shown in the workflow in Figure 1. The purpose of this design is to minimize the risk of misinterpretation by the model, especially if users provide a vague or incomplete prompt.

\begin{figure}[h]
  \centering
  \includegraphics[width=1\textwidth]{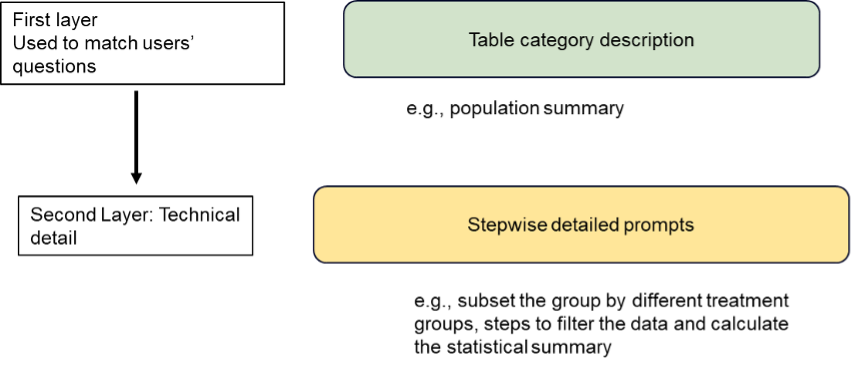}  
  \caption{Two-layer architecture of the Clinical Trial TFLs Generation Agent}
  \label{fig:your_label}
\end{figure}

The backend system matches users' requests with predefined table generation prompts. Here's how it works:
\begin{itemize}
    \item \textbf{Table Generation and Description Matching:} Each table has a detailed description associated with it. When a user makes a request, GPT-4 matches the user's question with the appropriate table description. If a match is found, the system will automatically run the predefined code generation prompts to create the table. The table description part needs close collaboration with statistician by describing the specific subgroup analysis in this table. 
    
    \item \textbf{Non-Code Generation Requests:} Not all user queries require code generation. For requests related to data background information or a general data overview, the app directly returns the relevant description without generating any code. To differentiate between code generation and non-code generation requests, we include a flag after each table description. This flag helps the model identify whether code generation is needed and process the request accordingly.
    
    \item \textbf{Unmatched Requests:}  If a user's request does not match any of the predefined table descriptions, the system will treat the request and their uploaded data as a new prompt, then it will be processed by GPT-4o to generate the answer, and we added features to allow users to choose if they want the code which used to generated the result, that is helpful for statisticians for future validation and adapt to their own analysis.
\end{itemize}

This approach ensures that users receive accurate and relevant responses based on their queries, streamlining the table generation process and providing comprehensive data insights efficiently. 

\section{Discussion and Future Direction}
In this project, we have investigated the feasibility of using a large language model to generate clinical trial TFL outputs, and explored a path to implement a user experience that can operationalize this in practice through a TFL Generation Agent application. Unlike general-purpose tools like OpenAI's ChatGPT, our Clinical Trial TFL Generation Agent is designed with a focus on clinical trial calculations, employing a prompt-based design that guides the LLM to produce consistent and accurate results using a standardized prompt library instead of an unguided chat interface. We ensured transparency of the computations performed by outputting generated program code along with the outputs. Our study showed that with careful design of prompts, LLMs be helpful as a tool to search a library of standard TFLs, as well as perform small study-specific adaptations to these prompts. Moreover, standardized prompts can accelerate previewing analysis results and help guide discussions between statisticians, programmers and clinical teams in a data-driven fashion while not increasing the risk of errors that may be introduced by imprecise specifications – and thus could enable question-driven data interrogation similar to Shiny applications\cite{teal_r_shiny_framework}.

Good performance for generating code and outputs heavily relies on the quality of the predefined prompts, as well as good quality and standards-compliance of the input data. Currently the model struggles with rules requiring complex mapping, coding or tasks requiring advanced/uncommon knowledge to implement the statistical analysis. To overcome this, we would need to develop a prompt library for the TFLs in the CSR that would capture this specialized knowledge, and could supplement existing code templates and macro libraries. We may also be able to improve performance by adding support for creating tables from standardized analysis results datasets\cite{cdisc_analysis_results_standard}, which would allow us to further separate programming tasks for data mapping and summarization from data visualization tasks. Additionally, the performance of larger models often improves with increased computational resources, and the raw data from the clinical trial can significantly impact the final results\cite{creswell2022selection,fan2024towards,sherry2023prevalence}.

Another potential enhancement to our Clinical Trial TFL Generation Agent is to assist with more complex computations. Enhancing the app to provide more precise support in these analyses would be highly beneficial – e.g. also capturing knowledge and pitfalls about method applications in the prompt (but also relying on statistical knowledge built into state of the art LLMs). Finally, capturing interactions between users and LLM-based applications is a good way to gather data to improve responses based on ongoing conversations. This continuous interaction will allow the app to adapt more effectively to user needs and provide increasingly accurate and relevant assistance. This is a much more direct feedback loop than what is typical in existing dashboard or analytical applications with pre-defined visualizations (e.g. R/Shiny) that are commonly used for exploratory data analysis.

In summary, we believe that LLMs can contribute substantially to making the process of TFL generation less manual and more standardized. We believe the key benefit lies in the ability of LLMs to retain a link between program code and natural language / plain text specifications, which allows them to naturally integrate with and enhance the existing processes for designing and producing TFL outputs. The key challenge is in careful prompt design and testing, such that correct outputs can be produced reliably.  


\vspace{-0.1in}
\paragraph{Acknowledgement}
This analysis was carried out when Yumeng Yang interned at Novartis. Yumeng Yang is a Predoctoral Fellow supported by the Cancer Prevention and Research Institute of Texas (CPRIT). We thank our collaborators for their contributions to this work. 

\paragraph{Conflict of interest }
Yumeng Yang has nothing to disclose. Gen Zhu, Peter Krusche, and Kristyn Pantoja report employment by and stock ownership in Novartis. 

\paragraph{Author Contributions}
All authors have reviewed and approved the analysis, contributed to the development and approval of the manuscript, and acknowledged the decision to submit the manuscript for publication.

\footnotesize
\bibliographystyle{vancouver}
\bibliography{References}

\end{document}